\documentclass[journal]{IEEEtranTIE}
\usepackage{amsmath,amsfonts}
\usepackage{algorithmicx}
\usepackage{array}
\usepackage[caption=false,font=normalsize,labelfont=sf,textfont=sf]{subfig}
\usepackage{textcomp}
\usepackage{stfloats}
\usepackage{url}
\usepackage{verbatim}
\usepackage{graphicx}
\usepackage{cite}
\usepackage{booktabs}
\usepackage{makecell}
\usepackage{amssymb}
\usepackage{algpseudocode}
\usepackage[ruled,norelsize,vlined,linesnumbered]{algorithm2e}
\usepackage{multirow}
\usepackage{picinpar}
\usepackage{flushend}
\usepackage[latin1]{inputenc}
\usepackage{colortbl}
\usepackage{soul}
\usepackage{pifont}
\usepackage{color}
\usepackage{alltt}
\usepackage[hidelinks]{hyperref}
\usepackage{enumerate}
\usepackage{siunitx}
\usepackage{breakurl}
\usepackage{epstopdf}
\usepackage{pbox}
\usepackage{fancyhdr}

\begin{document}

\definecolor{darkblue}{RGB}{0,0,205}
\definecolor{limegreen}{rgb}{0.2, 0.8, 0.2}
\definecolor{forestgreen}{rgb}{0.13, 0.55, 0.13}
\definecolor{greenhtml}{rgb}{0.0, 0.5, 0.0}

\makeatletter
\newcommand{\rmnum}[1]{\romannumeral #1}
\newcommand{\Rmnum}[1]{\expandafter\@slowromancap\romannumeral #1@}
\makeatother
\newcommand{\revise}[1]{\textcolor{darkblue}{#1}}

\title{A2VISR: An Active and Adaptive Ground-Aerial Localization System Using Visual Inertial and Single-Range Fusion}

\author{
	\vskip 1em
	
	Sijia Chen, Wei Dong

	\thanks{
	Sijia Chen and Wei Dong are with the State Key Laboratory of Mechanical System and Vibration, School of Mechanical Engineering, Shanghai Jiaotong University, Shanghai, 200240, China. Corresponding author: Wei Dong, E-mail: dr.dongwei@sjtu.edu.cn.
	}

	\thanks{
	Manuscript received Month xx, 2xxx; revised Month xx, xxxx; accepted Month x, xxxx.
	}
}

\maketitle
	
\begin{abstract}
It's a practical approach using the ground-aerial collaborative system to enhance the localization robustness of flying robots in cluttered environments, especially when visual sensors degrade. Conventional approaches estimate the flying robot's position using fixed cameras observing pre-attached markers, which could be constrained by limited distance and susceptible to capture failure. To address this issue, we improve the ground-aerial localization framework in a more comprehensive manner, which integrates active vision, single-ranging, inertial odometry, and optical flow. First, the designed active vision subsystem mounted on the ground vehicle can be dynamically rotated to detect and track infrared markers on the aerial robot, improving the field of view and the target recognition with a single camera. Meanwhile, the incorporation of single-ranging extends the feasible distance and enhances re-capture capability under visual degradation. During estimation, a dimension-reduced estimator fuses multi-source measurements based on polynomial approximation with an extended sliding window, balancing computational efficiency and redundancy. Considering different sensor fidelities, an adaptive sliding confidence evaluation algorithm is implemented to assess measurement quality and dynamically adjust the weighting parameters based on moving variance. Finally, extensive experiments under conditions such as smoke interference, illumination variation, obstacle occlusion, prolonged visual loss, and extended operating range demonstrate that the proposed approach achieves robust online localization, with an average root mean square error of approximately 0.09 m, while maintaining resilience to capture loss and sensor failures.
\end{abstract}


\markboth{IEEE TRANSACTIONS ON INDUSTRIAL ELECTRONICS}%
{}

\definecolor{limegreen}{rgb}{0.2, 0.8, 0.2}
\definecolor{forestgreen}{rgb}{0.13, 0.55, 0.13}
\definecolor{greenhtml}{rgb}{0.0, 0.5, 0.0}

\section{Introduction}

\IEEEPARstart{R}{ecently}, unmanned aerial vehicles (UAVs) have become a cost-effective solution for infrastructure inspections, particularly in challenging environments such as bridge undersides, interior tunnels, and large industrial facilities \cite{article,9900135,10876055}. Current maintenance procedures typically rely on pilot-assisted semi-autonomous modes, while there is a growing demand for fully autonomous \cite{drones7020089}. To meet this demand, developing robust localization methods that can adapt to environmental interference and dynamic changes is the priority.

Conventionally, there are two mainstream positioning approaches for autonomous flying robots. The first relies on external facilities, such as the Global Navigation Satellite System (GNSS) \cite{10745226}, motion capture systems (MCS) \cite{reNf2}, and fixed ultrawideband (UWB) frameworks \cite{shule_uwb-based_2020}. Although these methods offer high robustness, their dependence on pre-installed infrastructure and time-consuming calibration limits the dynamic adaptability in unknown environments. The second approach equips UAVs with onboard sensors, such as visual, optical, and ranging technologies \cite{nguyen_range-focused_2021}. While the onboard configuration enhances mobility, its robustness may be limited due to reliance on a single data source. In particular, under challenging environmental conditions, degraded perception fidelity could compromise the reliability of estimation \cite{11045071}. To address these issues, ground-aerial collaboration systems have emerged as a promising solution, combining the robustness of multi-source sensor fusion with the dynamic adaptability of mobile systems \cite{10472632, 10577156, 10891932}.

Currently, ground-aerial collaborative localization is commonly achieved using vision-based methods. Simultaneous Localization and Mapping (SLAM) can establish relative transformation between the ground and aerial robots by jointly processing their visual observations \cite{10582478}, but this typically demands high computational resources and considerable communication bandwidth. In contrast, Visual-Inertial Odometry (VIO) offers a more lightweight alternative by estimating motion from visual and inertial inputs through environmental feature tracking \cite{10342012}. However, due to the pre-integration process, VIO inevitably accumulates long-term drift. Moreover, under weak-texture or low-light conditions, even industrial-grade sensors such as the Intel RealSense T265 may experience degraded or failed VIO performance.

To cope with these issues, detection- and marker-based methods have been employed as alternative solutions. For instance, Xu et al. \cite{xu_decentralized_2020} proposed a decentralized visual-inertial-UWB fusion framework using YOLOv3-tiny to detect the shape of the aerial robot directly. While effective, YOLO-based methods require additional training and may degrade in harsh environments with fog, smoke or fluctuating illumination. Alternatively, marker-based approaches are conducted and can be categorized into passive and active methods. The passive approach involves attaching reflective artificial markers, such as ARTags \cite{7487146}, AprilTags \cite{11152325}, and ArUcos \cite{9765385}, to the aerial robot. However, these corner features may become unclear under low illumination \cite{Fiducial}. To overcome this issue, active reflective infrared (IR) markers are explored to improve observation \cite{10197620,8967660,7487248}. These IR markers differentiate from ambient light, ensuring reliable recognition in cluttered environments without retraining.

Although active IR marker-based methods can effectively adapt to extreme environments characterized by fluctuating illumination, two main challenges remain. First, the re-capture issue arises due to the infrared cameras' limited field of view (FOV). Occlusions or intermittent visibility losses can cause estimation divergence, resulting in the UAV remaining outside the camera's view and preventing re-capture. Regarding this issue, omnidirectional vision provides an intuitive solution. Attempts include fisheye cameras \cite{xu_omni-swarm_2022} and camera arrays \cite{11128305}, typically with stationary cameras. However, such methods come at the cost of additional mass and computational requirements, introducing new burdens. Moreover, although fisheye camera distortion can be corrected through additional processing, the resolution still varies from the center to the edge, and the rectified images generally contain fewer details compared with a perspective camera at the same distance.

Another challenge arises from the limited visual capture range, which often requires multiple ground vehicles to position aerial robots collaboratively over larger areas. However, such deployment reduces the scalability of marker-based methods. To realize wide-area localization with a single ground beacon, the fusion of inertial measurement units (IMUs) and ultra-wideband (UWB) technology \cite{10449450,10943167,sun_novel_2024} has emerged as a promising solution. By integrating non-visual sensors into ground-aerial collaborative systems, long-distance localization becomes feasible. Nguyen et al. \cite{9655461} present a resource-efficient visual-inertial-range framework that avoids loop closure and relies only on neighbor odometry. Additionally, Cao et al. \cite{cao_vir-slam_2021} present a system that leverages UWB ranging with one static anchor to correct the accumulated error whenever the anchor is visible. These practices have further enhanced estimation scalability. However, sensor performance in visual-based inertial and ranging fusion frameworks remains vulnerable to unpredictable environmental factors such as smoke interference, illumination changes, and obstacle occlusion. Severe visual intermittent may degrade the system to a single-anchor configuration, leading to local observability uncertainty and deteriorating estimation \cite{10184165}. To enhance robustness, it is essential to extend the horizon of historical estimates while dynamically evaluating measurement fidelity and adaptively adjusting sensor confidence \cite{9829196,10226597,8954658}.

To address the aforementioned challenges, we propose an active and adaptive ground-aerial localization framework that integrates active vision, single-ranging, inertial odometry, and optical flow. Specifically, the active vision subsystem, mounted on the ground vehicle with two servo motors allowing horizontal and vertical rotation, continuously detects and tracks the infrared markers on the aerial robot. This mechanism expands the field of view and improves the target recognition based on only a single camera. Besides, fusing single-ranging with inertial odometry extends the operational range and mitigates re-capture failures under visual degradation. Additionally, the aerial robot employs optical flow and a fixed-height laser to provide comprehensive velocity references. Based on these inputs, a dimension-reduced estimator is implemented with an extended sliding window that fuses multi-source measurements using polynomial approximation, balancing computational efficiency with redundancy retention. Considering the reliance on sensor feedback, an adaptive sliding confidence evaluation algorithm assesses measurement quality and dynamically adjusts the weighting of different terms based on moving variance. Built on this framework, the target aerial robot can be effectively positioned with only one cooperative ground robot, ensuring robustness even in extreme conditions.

The main contributions of this work are as follows: 1) An active and adaptive ground-aerial localization framework is proposed, integrating active infrared marker observation, single-range, inertial odometry and optical flow, enhancing the position robustness of the flying robot in harsh environments. 2) An augmented dimension-reduced estimator is reformulated, considering the dynamic assessment of sensor fidelities based on an adaptive sliding confidence evaluation algorithm.

\section{Notation and Problem Formulation}
\begin{figure}[!t]\centering
	\includegraphics[width=8.5cm]{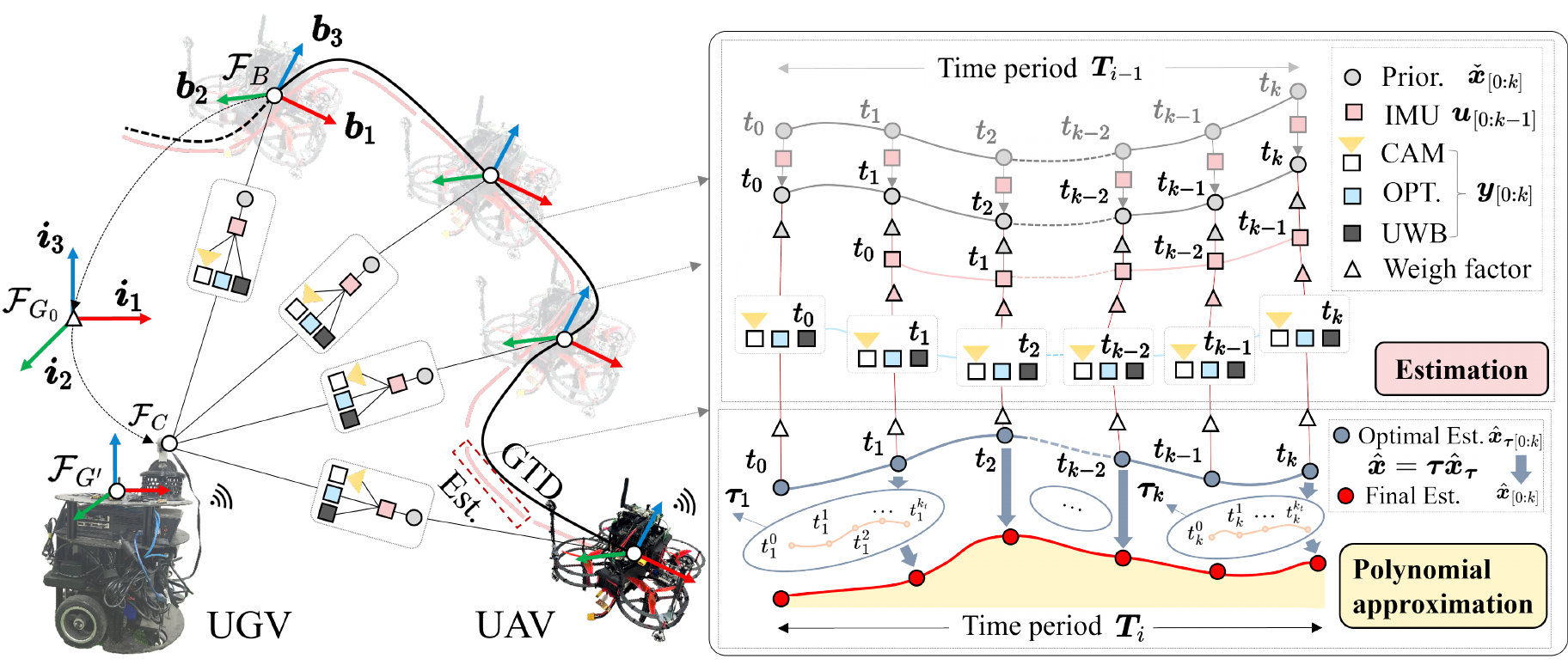}
	\caption{The ground-aerial localization framework. (For an extended period, the prior estimates, active visual feedback, IMU, optical, and distance measurements are acquired. Subsequently, an augmented dimension-reduced estimator is reformulated to perform polynomial approximation.)}\label{framework}
\end{figure}

To achieve robust localization, we present a ground-aerial cooperative system comprising a ground vehicle as the monitor and an aerial robot as the target. The estimation framework is illustrated in Fig. \ref{framework}. In this section, we establish the notation and provide the preliminary problem formulation.

\subsection{Notation}
In this work, a matrix with dimension $m$ by $n$ is denoted by a bold capital letter as $\boldsymbol {M} \in \mathbb{R}^{m \times n}$. The vector with dimension $n$ is denoted by a bold lowercase letter, $\boldsymbol{x} \in \mathbb{R}^{n}$. For $\boldsymbol{x} \in \mathbb{R}^{n}$ and $\boldsymbol {M} \in \mathbb{R}^{n \times n}$, we define the norm $\|\boldsymbol{x}\|_{\boldsymbol{M}}=\boldsymbol{x}^{\top} \boldsymbol{M} \boldsymbol{x}$. The identity and zero matrices are denoted as $\boldsymbol{I}_{m \times n}$ and $\boldsymbol{O}_{m \times n}$, respectively; their square matrices of dimension $n$ are abbreviated as $\boldsymbol{I}_n$ and $\boldsymbol{O}_n$. For a matrix, $(\cdot)^{\top}$ denotes its transpose and $(\cdot)^{-1}$ denotes the inverse. For a vector, $\left\|\cdot\right\|_2$ represents for its Euclidean norm. The notation $\mathrm{diag}(\boldsymbol{X}_{n}, \dots, \boldsymbol{Y}_{m})$ refers to a block diagonal matrix. To distinguish prior and posterior estimates, we use the breve description $\check{(\cdot)}$ and $\hat{(\cdot)}$, respectively. During estimation, the width of a sliding window is defined as $T_w$. The state sequence within interval $k$ is indicated as $\left[{\boldsymbol{x}}_k\right]_{0}^{T_w} = [{\boldsymbol{x}_0, \boldsymbol{x}_1, \dots, \boldsymbol{x}_{T_w}}]_k$. Accordingly, the posterior state sequence $\left[\hat{\boldsymbol{x}}_{k-1}\right]_{0}^{T_w}$ in interval $k-1$ corresponds to the prior sequence $\left[\check{\boldsymbol{x}}_k\right]_{-1}^{T_w-1}$ in the subsequent iteration $k$.

\begin{figure}[!t]\centering
	\includegraphics[width=8.5cm]{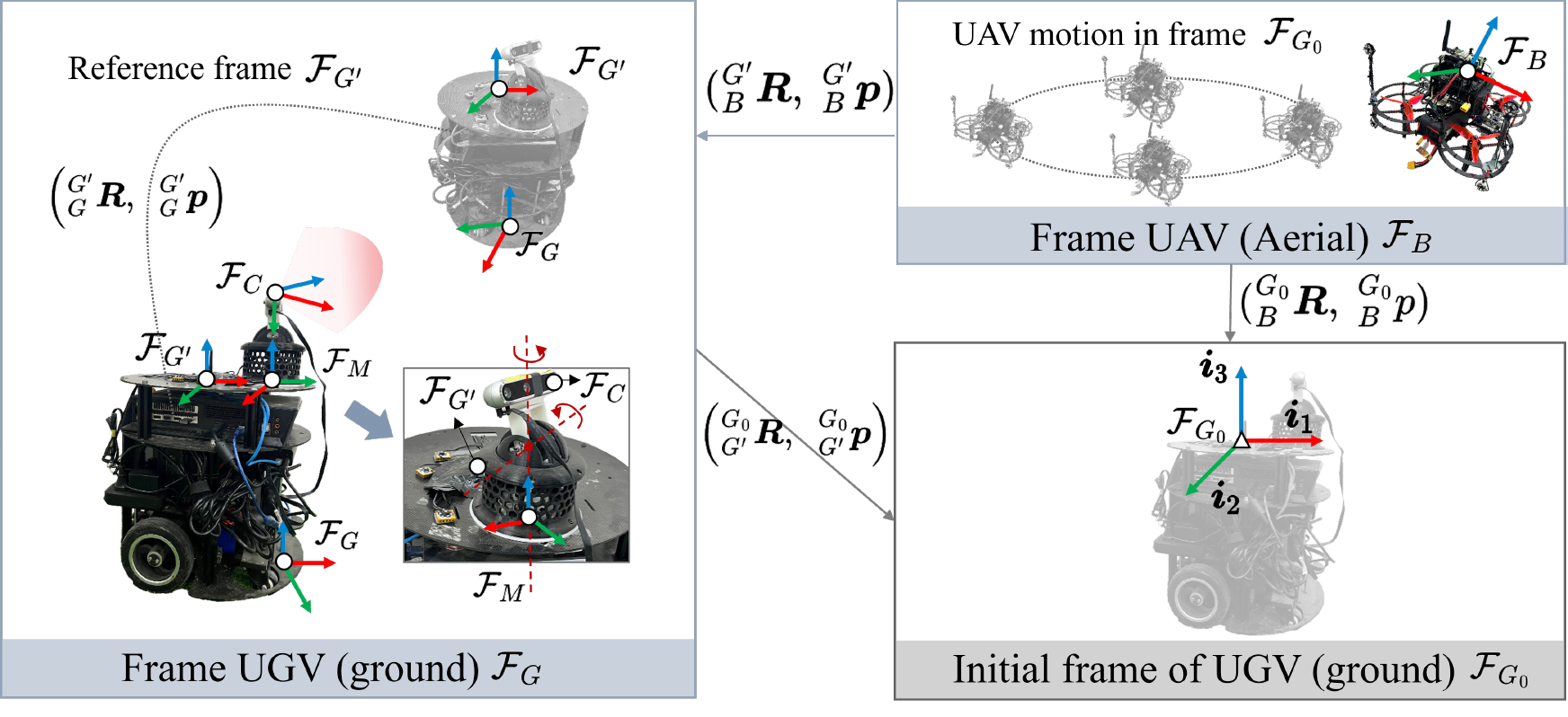}
	\caption{Coordinate transformations are defined among the aerial robot (body frame $\mathcal{F}_{B}$), the ground vehicle (ground frame $\mathcal{F}_{G}$), and the ground vehicle's initial frame $\mathcal{F}_{G_0}$. The ground vehicle and the aerial robot are controlled using reference commands expressed in the initial frame.}\label{Coordinate}
\end{figure}

The coordinates are represented by capital calligraphic letters, and the transformations are illustrated in Fig. \ref{Coordinate}. Specifically, $\mathcal{F}_{B}$ denotes the body frame of the aerial robot, while $\mathcal{F}_{G}$ represents the ground frame attached to the ground vehicle. The frame $\mathcal{F}_{G_0}$ serves as the initial pose of $\mathcal{F}_{G}$, analogous to the inertial global frame. To decouple the robots' motion, we define a dynamic reference frame $\mathcal{F}_{G^{\prime}}$ based on the inertial frame $\mathcal{F}_{G_0}$. As the ground vehicle moves and rotates, $\mathcal{F}_{G^{\prime}}$ inherits only the translational degrees of freedom from $\mathcal{F}_{G}$, maintaining synchronized displacement while keeping its initial orientation fixed. The frame $\mathcal{F}_{C}$ corresponds to the camera, and $\mathcal{F}_{M}$ represents the frame at the base of the active vision mechanism. The rotation matrix ${}^{A}_{B} \boldsymbol{R} \in \mathrm{SO}(3)$ defines the transformation from $\mathcal{F}_{B}$ to $\mathcal{F}_{A}$. Physical vectors expressed in their respective coordinates are indicated by the left superscripts. For example, the relative position of $\mathcal{F}_{B}$ with respect to $\mathcal{F}_{G}$ is denoted as ${}^{G}_{B} \boldsymbol{p}$, while the relative position of $\mathcal{F}_{G}$ with respect to itself is written as ${}^{G} \boldsymbol{p}$.

\subsection{Problem Formulation}
In the ground-aerial cooperative system, multi-source fusion incorporates acceleration, distance, and optical velocity measurements from the aerial robot, while the ground vehicle handles reference visual tracking. To achieve precise localization, the problem is initially formulated using the Maximum A Posteriori (MAP) approach as: $\hat{\boldsymbol{x}}=\arg \max _{{\boldsymbol{x}}} p(\boldsymbol{x}|\check{\boldsymbol{x}},\boldsymbol{u}, \boldsymbol{y})$, where $\check{\boldsymbol{x}}$, $\boldsymbol{u}$, $\boldsymbol{y}$ represent for the prior, input, and measurements respectively. By applying Bayes' theorem and assuming independent process and measurement noise with invertible covariances \cite{refN8}, the MAP formulation is transformed into an optimization problem: $\hat{\boldsymbol{x}}=\arg \min _{{\boldsymbol{x}}} \boldsymbol{J}$. Given the sliding window width $T_w$, the objective function at timestep $k$ consists of $N$ terms, then the quadratic cost function for estimating the optimal state sequence is formulated as:
\begin{equation}
  \boldsymbol{J}= \sum_{i=1}^{N} \sum_{k=0}^{T_w}\left\|\mathcal{E}_i({\boldsymbol{x}}_k)-\mathcal{E}_i(\tilde{\boldsymbol{x}}_k) \right\|_{{}^i\boldsymbol{W}_{k}}
  \label{eq:optimalization}
\end{equation}
\noindent where ${}^i\boldsymbol{e}_k=\mathcal{E}_i({\boldsymbol{x}}_k)-\mathcal{E}_i(\tilde{\boldsymbol{x}}_k)$ denotes the influential error. The weighted norm is defined as $\|{}^i\boldsymbol{e}_k\|_{{}^i\boldsymbol{W}_{k}}={}^i\boldsymbol{e}_k^{\top} {}^i\boldsymbol{W}_{k} {}^i\boldsymbol{e}_k$, where ${}^i\boldsymbol{W}_{k}$ is the confidence evaluation matrix for the $i$-th sensor. $\mathcal{E}_i({\boldsymbol{x}}_k)$ represents its observation model, and the $\mathcal{E}_i(\tilde{\boldsymbol{x}}_k)$ denotes the general description of measurements, incorporating the prior states, control input, and sensor feedback.

\section{Active and Adaptive Estimation Based on Visual Inertial and Single-Range Fusion}
In this section, we present the active and adaptive estimation process, integrating visual inertial and single-range fusion. Initially, we address coordinate transformations to eliminate dependencies on the inertial world frame. Then, we introduce the active vision mechanism mounted on the ground vehicle, which provides reference active visual tracking for the aerial robot. Subsequently, we detail the adaptive sliding confidence evaluation process, which assesses the quality of measurements. Finally, we extend the augmented dimension-reduced estimator to mitigate computational costs.

\subsection{Dynamics and Coordinate Transformations}
In our work, taking the initial frame $\mathcal{F}_{G_0}$ as the intermediate, we can derive the dynamics of the aerial robot transferred from body coordinates $\mathcal{F}_{B}$ to the dynamic reference frame of the unmanned ground vehicle $\mathcal{F}_{G^{\prime}}$ as follows.
\begin{equation}
  \begin{aligned}
  {}^{G^{\prime}}_{B}{\boldsymbol{p}} = {}^{G^{\prime}}_{G_0}\boldsymbol{R} ({}^{G_0}_{B}\boldsymbol{p} - {}^{G_0}_{G^{\prime}}\boldsymbol{p}) \\
  \end{aligned}
  \label{dynamic_pos}
\end{equation}
\noindent Since $\mathcal{F}_{G^{\prime}}$ undergoes only translational motion relative to $\mathcal{F}_{G_0}$, the rotation matrix ${}^{G^{\prime}}_{G_0} \boldsymbol{R}$ equals to the identity matrix. ${}^{G_0}_{B}\boldsymbol{p}=x_b \boldsymbol{i}_1+y_b \boldsymbol{i}_2+z_b \boldsymbol{i}_3$ represents the relative position of the aerial robot's frame $\mathcal{F}_{B}$ with respect to the initial ground frame $\mathcal{F}_{G_0}$. Similarly, the relative position of the the reference ground frame $\mathcal{F}_{G^{\prime}}$ with respect to $\mathcal{F}_{G_0}$ can be expressed as ${}^{G_0}_{G^{\prime}}\boldsymbol{p}=x_g \boldsymbol{i}_1+y_g \boldsymbol{i}_2+z_g \boldsymbol{i}_3$, which can be obtained through the external positioning device. Here, $\boldsymbol{i}_n (n=1,2,3)$ denotes the unit vector in the initial frame.

In practical scenarios, the relative position of $\mathcal{F}_{G^{\prime}}$ is first estimated, after which the position in the initial frame ${}^{G_0}_{B}\boldsymbol{p}$ can be derived through (\ref{dynamic_pos}) and used to control the motion of both the aerial and ground vehicles. Besides, requiring the ground vehicle to move at a constant speed and turns with a constant angular velocity, ${}^{G_0}_{B}{\dot{\boldsymbol{v}}}$ can be approximated as zero. The nominal-state system model can be formulated as.
\begin{equation}
  \begin{aligned}
    {}^{G^{\prime}}_{B}{\dot{\boldsymbol{p}}} & = {}^{G_0}_{B}{\boldsymbol{v}} - {}^{G_0}_{G^{\prime}}{\boldsymbol{v}} = {}^{G_0}_{B}\boldsymbol{R} \cdot {}^{B}{\boldsymbol{v}} - {}^{G_0}_{G^{\prime}}{\boldsymbol{v}}\\
    {}^{G^{\prime}}_{B}{\dot{\boldsymbol{v}}} & = {}^{G_0}_{B} {\boldsymbol{u}} -\boldsymbol{\mu}\cdot {}^{G_0}_{B}\boldsymbol{v}\\
  \end{aligned}
  \label{eq:dynamic_vel}
\end{equation}
\noindent where ${}^{B} {\boldsymbol{v}}$ denotes the velocity for the aerial robot obtained in the body frame. Since the raw acceleration measurements collected from IMU are normalized, the actual acceleration input can be obtained by: ${}^{G_0}_{B}{\boldsymbol{u}} = g \cdot {}^{G_0}_{B}\boldsymbol{R} \cdot {}^{B}{\boldsymbol{a}} - [0,0,g]^{\top}$, and $g$ denotes the gravity constant. ${}^{B}{\boldsymbol{a}} = [a^x,a^y,a^z]^{\top}$ denotes the linear acceleration collected from IMU. ${}^{B}{\boldsymbol{q}} = [q^w,q^x,q^y,q^z]^{\top}$ denotes the quaternion referring to its initial state. During initialization, ensure that the initial quaternion of the aerial robot and the ground vehicle are consistent. Then ${}^{G_0}_{B}\boldsymbol{R}$ can be obtained through ${}^{B}{\boldsymbol{q}}$. Meanwhile, we additionally consider the linear drag effect ${}^{G_0}_{B} \boldsymbol f_{\mu}=-\boldsymbol{\mu} m {}^{G_0}_{B}\boldsymbol{v}$, which inherently reflects the dissipative nature of UAV in real world.

For simplicity, we write the relative quantities ${}^{G^{\prime}}_{B}{\boldsymbol{p}}$ and ${}^{G^{\prime}}_{B}{\boldsymbol{v}}$ as $\boldsymbol{p}$, $\boldsymbol{v}$ respectively. The relative state is denoted as $\boldsymbol{x}^{\top} = \left[\boldsymbol{p}^{\top}, \boldsymbol{v}^{\top}\right]$. The superscripts will be omitted in the following description. By pre-integration with a sampling frequency $\mathrm{d}t$, the discrete state function can be formulated and linearized as:
\begin{equation}
  \begin{aligned}
  \boldsymbol{x}_{k+1} & =\left[\begin{array}{ll}
  \boldsymbol{I}_3 & \mathrm{~d} t \boldsymbol{I}_3 \\
  \mathbf{0}_3 & \boldsymbol{I}_3-\mathrm{d} t \boldsymbol{\mu}_k
  \end{array}\right] \boldsymbol{x}_k+\left[\begin{array}{c}
  \frac{1}{2} \mathrm{~d} t^2 \\
  \mathrm{~d} t
  \end{array}\right] \otimes \boldsymbol{I}_3 \boldsymbol{u}_k \\
  & =\boldsymbol{A}_k \boldsymbol{x}_k+\boldsymbol{B}_k \boldsymbol{u}_k
  \end{aligned}
  \label{eq:discrete model}
\end{equation}
\noindent where ${\boldsymbol{x}} \in \mathbb{R}^{6}$ denotes the relative state vector, $\boldsymbol{u} \in \mathbb{R}^{3}$ denotes the input vector corresponding to acceleration, and $\boldsymbol{\mu}_k \in \mathbb{R}^{3 \times 3}$ denotes the aerial drag coefficient matrix, $\boldsymbol{\mu}_k = \mathrm{diag}(\mu^x_k,\mu^y_k,\mu^z_k)$. $\boldsymbol{A}_{k}$ and $\boldsymbol{B}_{k}$ represent the system matrix and input matrix, respectively. $\otimes$ represents the Kronecker product.

As for system function, the output comprises distance, velocity, and reference visual feedback. First, we approximately reformulate the nonlinear distance measurement obtained from the UWB to a linearized form based on prior feedback:
\begin{equation}
  {}^{\mathrm{UWB}}y_k=\left[\begin{array}{ll}
    \boldsymbol{\rho}_k^{\top} & \mathbf{0}_{1 \times 3}
  \end{array}\right] {\boldsymbol{x}}_k
  \label{eq:uwb_observation}
\end{equation}
\noindent where $\boldsymbol{\rho}_k=\boldsymbol{r}^{\top}_{k} /\left\|\boldsymbol{r}_{k}\right\|_2$. The $\boldsymbol{r}_{k}$ is an approximate position calculated via $\boldsymbol{r}_{k}=\left[\begin{array}{ll}
\boldsymbol{I}_3 & \boldsymbol{0}_3
\end{array}\right]\left(\boldsymbol{A}_{k-1} \check{\boldsymbol{x}}_{k-1}+\boldsymbol{B}_{k-1} \boldsymbol{u}_{k-1}\right)$.

During flight, the aerial robot is equipped with an optical flow sensor and a laser altimeter to measure velocity ${}^{B}\boldsymbol{v}_k$ and relative height ${}^{G_0}_{B} h_k$. The relative height is expressed as ${}^{\mathrm{ALT}}{y}_k = {}^{G_0}_{B} h_k - h_g$, where $h_g$ presents a fixed height in $\mathcal{F}_{G_0}$ relative to take-off plane. The optical flow provides velocity components along the $x$ and $y$ axes, while the $z$-axis velocity is inferred from the laser altimeter. Since the $z$-axis direction of $\mathcal{F}_B$ remains relatively constant and the velocity ${}^{G_0}_{G^{\prime}}\boldsymbol{v}$ of ground vehicle can be obtained from wheel encoder. The relative velocity can be transformed based on equation (\ref{eq:dynamic_vel}): ${}^{\mathrm{OPT}}\boldsymbol{y}_k = {}^{G_0}_{B} \boldsymbol{R} [{}^{B}\boldsymbol{v}_k^{\top} ,({}^{G_0}_{B} h_k-{}^{G_0}_{B} h_{k-1})/\mathrm{dt}]^{\top}- {}^{G_0}_{G^{\prime}}\boldsymbol{v}_k$.

Additionally, reference visual tracking feedback from the ground vehicle provides position measurements. The detailed transition is presented in Section \uppercase\expandafter{\romannumeral3}-B. The observation model is then formulated as follows:
\begin{equation}
  \begin{aligned}
    \boldsymbol{y}_k&=\left[\begin{array}{cccc}
      {}^{\mathrm{UWB}}y_k^{\top} & {}^{\mathrm{OPT}}\boldsymbol{y}^{\top}_k &
      {}^{\mathrm{ALT}}y_k^{\top} & {}^{\mathrm{CAM}}\boldsymbol{y}^{\top}_k
      \end{array}\right]^{\top}\\
      &=\left[\begin{array}{cccc}
      \boldsymbol{\rho}_k & \boldsymbol{0}_{3} & 
      \boldsymbol{\beta}_k & \boldsymbol{I}_3 \\
      \boldsymbol{0}_{3 \times 1} & \boldsymbol{I}_3 &
      \boldsymbol{0}_{3 \times 1} & \boldsymbol{0}_3
      \end{array}\right]^{\top} \boldsymbol{x}_k =\boldsymbol{C}_{k} \boldsymbol{x}_k
  \end{aligned}
  \label{eq:observation}
\end{equation}
\noindent where $\boldsymbol{\beta}_k = [0,0,1]^{\top}$, $\boldsymbol{C}_{k}$ denotes measurement matrix.

In this system, the measurement feedback comprises comprehensive position and velocity information. The distance constraint mitigates potential divergence caused by long-term inertial integration, illumination degradation, and detection failures beyond the feasible range. Optical flow eliminates operational range limitations, while visual tracking provides high-precision estimates, particularly in small-scale scenarios. By integrating visual and non-visual sensors, the system combines the advantages of near- and far-field operation, ensuring robustness and adaptability in harsh environments.

\subsection{Reference Visual Estimation through Active Tracking}
The proposed active vision mechanism consists of two servo motors, realizing $360^{\circ}$ horizontal omnidirectional coverage and a $\pm 90^{\circ}$ pitch range for wide-area observation. To address challenging conditions such as low light and fluctuating illumination, the aerial robot is equipped with a rectangular array of four infrared markers (4-IR markers) for visual enhancement. An infrared filter is applied for the ground camera to distinguish the glowing infrared markers from the natural features.

Initially, the transformation ${}^{G_0}_{M_0}\boldsymbol{T}$ between the initial frame $\mathcal{F}_{G_0}$ and the initial base frame of the mechanism $\mathcal{F}_{M_0}$ will be established, and the motor angles will be initialized. During raw visual estimation, the captured image undergoes binarization, followed by a feature selection strategy based on geometric constraints (including parallel, perpendicular, and left-right analysis) to identify and prioritize the best four landmarks. Once a valid set of markers is detected, a perspective-n-point (PnP) algorithm is applied to compute the relative pose ${}^{C}_{B}\boldsymbol{T}$ of the target in the camera frame. Simultaneously, the joint angles from the servo motor encoder $(\theta,\phi)$ are acquired to calculate the transformation ${}^{M}_{C}\boldsymbol{T}$ from the camera frame $\mathcal{F}_C$ to the base of the mechanism frame $\mathcal{F}_M$. The relative pose is then determined through the composite transformation chain: ${}^{G^{\prime}}_{B}\boldsymbol{T} = {}^{G^{\prime}}_{G}\boldsymbol{T}{}^{G_0}_{M_0}\boldsymbol{T} {}^{M}_{C}\boldsymbol{T} {}^{C}_{B}\boldsymbol{T}$, where ${}^{G^{\prime}}_{G}\boldsymbol{T}$ is obtained through external localization feedback. Finally, the reference visual feedback is synchronized to the UAV.

During this process, joint angles $(\theta, \phi)$ are derived through inverse kinematics based on the fusion estimation at the last timestamp synchronized from the UAV, ensuring the projection of the target onto the camera plane center for tracking. Even if the landmarks are out of visible range, the active vision system maintains tracking relying on the current fusion, ensuring continuity for visual re-capture.

\subsection{Adaptive Sliding Confidence Evaluation}
Due to differences in measurement mechanisms, onboard sensors may experience degradation under unpredictable environmental changes. Additionally, variations in noise distribution among sensors of the same model can lead to inconsistent outcomes. These unmodeled, time-varying disturbances significantly impact the accuracy of multiple sensor fusion for positioning. Yang et al. \cite{yang_resilient_2021} proposed a resilient approach that switches positioning strategies based on an assessment of sensor interference. However, when measurements remain frozen for an extended period, this method relies solely on the last available data, potentially leading to divergence. To address this issue, we propose an adaptive sliding confidence evaluation algorithm. First, failure assessment is performed based on feedback from sensor measurement variations.
\begin{equation}
  {}^i\boldsymbol{S}_{f, k}= 
  \begin{cases}\boldsymbol{I}_s & \min \left({}^i\boldsymbol{\omega}_f\right)>\epsilon_f \\
    \varepsilon \boldsymbol{I}_s & \min \left({}^i\boldsymbol{\omega}_f\right)\leq \epsilon_f
  \end{cases}\label{eq:S failure}
\end{equation}
\noindent where the status ${}^i\boldsymbol{S}_{f, k}$ the failure condition of the $i$-th sensor at timestamp $k$. The identity matrix $\boldsymbol{I}_s$ represents a valid sensor state, and $s$ denotes the dimension of the sensor feedback. $\varepsilon$ is a small constant used to avoid numerical errors. The sensor residual is computed as ${}^i\boldsymbol{\omega}_f = \sum_{k=0}^{T_w}|{}^i\boldsymbol{y}_k - {}^i\boldsymbol{y}_{k-1}| \in \mathbb{R}^s$. Each element corresponds to the accumulated residual along one axis. The minimum component of this vector is compared with the threshold $\epsilon_f$ to determine long-term sensor failure during window size $T_w$. If $\min({}^i\boldsymbol{\omega}_f) \le \epsilon_f$, ${}^i\boldsymbol{S}_{f,k}$ is set to $\varepsilon \boldsymbol{I}_s$, reflecting long time lost. Thus, the corresponding sensor is considered to be invalid for this period. Otherwise, it is considered operating normally.

Meanwhile, the measurement quality is evaluated as:
\begin{equation}
{}^i \boldsymbol{S}_{q, k} = \boldsymbol{I}_s - \operatorname{diag}\left(\sigma({}^i\boldsymbol{\omega}_q)\right)
\label{eq:S quality}
\end{equation}
\noindent where ${}^i\boldsymbol {\omega}_q = |{}^i\boldsymbol{y}_{k} - {}^i\boldsymbol{y}_{k-1}|$ denotes the element-wise absolute difference between consecutive measurements. The sigmoid function $\sigma(\cdot)$ is applied element-wise as: $\sigma(({}^i\boldsymbol{\omega}_q)_j) = 1/(1+ e^{-m(({}^i\boldsymbol{\omega}_q)_j -{\omega}_0)}),j=1,\cdots,s$. The scalar coefficients $m$ and $\omega_0$ are preset. Accordingly, more pronounced outliers correspond to lower quality feedback.

In order to account for performance fluctuations during motion, the sensor confidence is dynamically indicated. During evaluation, the position $\overline{\boldsymbol{x}}_k$ from the trajectory planner serves as reference, and the moving variance between the reference and measured states is computed within each cycle $T_w$.
\begin{equation}
  \left\{\begin{aligned}
    { }^1 \boldsymbol{P}_k & =\sum_{k=0}^{T_w} \mathbb{E}\left[\left(\overline{\boldsymbol{x}}_k-\check{\boldsymbol{x}}_k\right)\left(\overline{\boldsymbol{x}}_k-\check{\boldsymbol{x}}_k\right)^{\top}\right] \\
    { }^i \boldsymbol{P}_k &=\sum_{k=0}^{T_w}\mathbb{E}\left[\left({}^i\boldsymbol{C}_k \overline{\boldsymbol{x}}_k-{ }^i \boldsymbol{y}_k\right)\left({}^i\boldsymbol{C}_k \overline{\boldsymbol{x}}_k-{ }^i \boldsymbol{y}_k\right)^{\top}\right]\\
    \end{aligned}\right.
  \label{eq:S variation}
\end{equation}
\noindent where ${ }^i \boldsymbol{P}_k (i = 1,2,\cdots,n_l)$ presents the moving variance for the inertial, distance, altimeter, optical flow, and visual measurements. $n_l$ represents the number of sensors. And the prior is given by $\check{\boldsymbol{x}}_k = \boldsymbol{A}_{k-1} \check{\boldsymbol{x}}_{k-1}+\boldsymbol{B}_{k-1} \boldsymbol{u}_{k-1}$. Then, the adaptive update of ${ }^i {\gamma}_k^{(d)}$ is formulated as ${ }^i {\gamma}_k^{(d)}=1-{ }^i \boldsymbol{P}_{k,d}/\sum_{j=1}^{n_l}\mathrm{tr}({ }^j \boldsymbol{P}_{k})$. Here, ${ }^i {\gamma}_k^{(d)}$ denotes the normalized weight for the $d$-th element of the $i$-th measurement, while ${ }^i\boldsymbol{P}_{k,d}$ represents the $d$-th diagonal element in matrix ${ }^i\boldsymbol{P}_{k}$. The normalized weight matrix is obtained by ${ }^i \boldsymbol{\gamma}_k=\operatorname{diag}\left({ }^i \gamma_k^{(1)},{ }^i \gamma_k^{(2)}, \ldots,{ }^i \gamma_k^{(s)}\right)$. Notably, a smaller moving variance in a specific dimension (i.e., a smaller diagonal element) results in a larger ${ }^i \boldsymbol{\gamma}_k^{(d)}$, indicating higher reliability.

Finally, the weights matrix is adaptively updated per sensor and per measurement axis according to failure feedback, quality assessment, and moving variance. This reduces the effect of unmodeled disturbances and noise.
\begin{equation}
  { }^i \boldsymbol{W}_k=\Xi \times \frac{{ }^i \boldsymbol{S}_{f, k} \odot{ }^i \boldsymbol{S}_{q, k} \odot{ }^i \boldsymbol{\gamma}_k}{\sum_{j=1}^{n_l} \operatorname{tr}\left({ }^j \boldsymbol{S}_{f, k} \odot{ }^j \boldsymbol{S}_{q, k} \odot{ }^j \boldsymbol{\gamma}_k\right)}
  \label{eq:S weighing}
\end{equation}
\noindent where $\Xi$ is the sum of weights for distribution, and ${ }^i \boldsymbol{W}_k (i=1,2,\dots,n_l, n_l =5)$ refer to the weighing matrix to the inertial, distance, altimeter, optical flow, and reference visual components, respectively. $\odot$ denotes the Hadamard Product.

\subsection{Augmented Dimension-Reduced Estimator}
In this section, supplementary measurements are integrated with an extended sliding window framework to improve long-term observability. As described in Section \uppercase\expandafter{\romannumeral2}-B, the objective function can be preliminarily designed as:
\begin{equation}
  \begin{aligned}
    \boldsymbol{J}= & \sum_{k=0}^{T_w}\left\|\hat{\boldsymbol{x}}_k-\check{\boldsymbol{x}}_k\right\|_{{}^{p}\boldsymbol{W}_k} +\sum_{k=1}^{T_w}\left\|\hat{\boldsymbol{x}}_k-\tilde{\boldsymbol{x}}_k\right\|_{{}^{1}\boldsymbol{W}_k}\\
    & +\sum_{i=2}^{n_l}\sum_{k=0}^{T_w}\left\|^i \hat{\boldsymbol{y}}_k-{ }^i \boldsymbol{y}_k\right\|_{{}^{i}\boldsymbol{W}_k}
    \end{aligned}
  \label{eq:optimalization real}
\end{equation}
\noindent where $\check{\boldsymbol{x}}_k = \boldsymbol{A}_{k-1} \check{\boldsymbol{x}}_{k-1}+\boldsymbol{B}_{k-1} \boldsymbol{u}_{k-1}$, $\tilde{\boldsymbol{x}}_k = \boldsymbol{A}_{k-1} \hat{\boldsymbol{x}}_{k-1}+\boldsymbol{B}_{k-1} \boldsymbol{u}_{k-1}$, and $^i\hat{\boldsymbol{y}}_k ={}^i\boldsymbol{C}_k \hat{\boldsymbol{x}}_{k}$. The weighing matrices ${}^{p}\boldsymbol{W}_k$, ${}^{1}\boldsymbol{W}_k$, ${}^{i}\boldsymbol{W}_k (i=2,3,\cdots,n_l)$ correspond to the covariance matrices of prior estimation, state transfer and measurement.

For simplicity, the maximum a posterior estimation is expressed as $\hat{\boldsymbol{x}}=\arg \min_{\boldsymbol{x}} J=\boldsymbol{E}^{\mathrm{T}} \boldsymbol{W} \boldsymbol{E}$, where the error matrix $\boldsymbol{E}$ is rewritten as:
\begin{equation}
  \boldsymbol{E}=\left[\begin{array}{c}
  \boldsymbol{I}_{n_x} \\
  \tilde{\boldsymbol{A}} \\
  \tilde{\boldsymbol{C}}
  \end{array}\right] \hat{\boldsymbol{x}}-\left[\begin{array}{ccc}
  \boldsymbol{I}_{n_x} & \boldsymbol{0} & \boldsymbol{0} \\
  \boldsymbol{0} & \tilde{\boldsymbol{B}} & \boldsymbol{0} \\
  \boldsymbol{0} & \boldsymbol{0} & \boldsymbol{I}_{n_y}
  \end{array}\right] \boldsymbol{\alpha} \triangleq \boldsymbol{E}_{\boldsymbol{x}} \hat{\boldsymbol{x}}-\boldsymbol{E}_{\boldsymbol{\alpha}} \boldsymbol{\alpha}
  \label{eq:estimation process}
\end{equation}
\noindent where $n_x = 6(T_w+1)$ and $n_y = 8(T_w+1)$ relative to the dimension of prior estimates and measurements. The block diagonal matrix $\tilde{\boldsymbol{A}} = \bigoplus_{k=1}^{T_w} \tilde{\boldsymbol{A}}_k$ is composed of $\tilde{\boldsymbol{A}}_k = \left[ -\boldsymbol{A}_k, \boldsymbol{I}_6 \right]$. Similarly we use the direct sum $\oplus$ to simplify notation, representing $T_w$ copies of matrix $\boldsymbol{B}_k$ along the diagonal as $\tilde{\boldsymbol{B}}=\bigoplus_{k=1}^{T_w}\boldsymbol{B}_k$. $\tilde{\boldsymbol{C}}=\bigoplus_{k=0}^{T_w}\boldsymbol{C}_k$ aggregates the measurement matrices. The posterior estimates over current window are represented by $\hat{\boldsymbol{x}} = [\hat{\boldsymbol{x}}_0^{\top}, \hat{\boldsymbol{x}}_1^{\top}, \cdots, \hat{\boldsymbol{x}}_{T_w}^{\top}]^{\top}$, with the shorthand notation $\hat{\boldsymbol{x}} = \left[\hat{\boldsymbol{x}}_k\right]_{0}^{T_w}$ to simplify the concatenated column vector. Similarly, $\boldsymbol{\alpha} = [\left[\check{\boldsymbol{x}}_k\right]_{0}^{T_w},\left[{\boldsymbol{u}}_k\right]_{1}^{T_w},\boldsymbol{Y}]$ concatenates prior estimates, control inputs, and supplementary measurements, where $\boldsymbol{Y}=\left[{}^i\boldsymbol{Y}\right]_{i=1}^4$ and ${}^i\boldsymbol{Y}=\left[{ }^i \boldsymbol{y}_k\right]_{0}^{T_w}$. The weighting matrix is given by $\boldsymbol W = \bigoplus_{k=0}^{T_w} {}^p\boldsymbol{W}_k \oplus \bigoplus_{k=1}^{T_w} {}^1\boldsymbol{W}_k \oplus \bigoplus_{k=0}^{T_w} {}^r\boldsymbol{W}_k$, where ${}^r\boldsymbol{W}_k =  \bigoplus_{i=1}^{4} {}^i\boldsymbol{W}_k$.

\begin{algorithm}[t]
  \caption{ Extended sliding window filter considering dimension reduced process with adaptive confidence evaluation at timestamp $k$.}
	\label{alg:algorithm1}
  \SetAlgoLined
	\KwIn{$\overline{\boldsymbol{x}}_k$, $\check{\boldsymbol{x}}_k$, $\tilde{\boldsymbol{x}}_k$, $\boldsymbol{u}_k$, ${ }^i \boldsymbol{y}_k$, $\boldsymbol{A}_k$, $\boldsymbol{B}_k$, ${}^i\boldsymbol{C}_k$,\\
  \qquad \quad $T_w$, $k$, $\mathrm{d}t$, $n_l$, $k_t$, $m$, $\omega_0$, $\Xi$, $\varepsilon$, $\epsilon_f$, $\boldsymbol{\mu}$, ${ }^p \boldsymbol{W}_k$}
	\KwOut{The optimal posterior states $\hat{\boldsymbol{x}}_k$}
	\BlankLine
  \For{$i=1$ \KwTo $n_l$}{
    Calculate ${}^i\boldsymbol{S}_{f, k}$ and ${}^i \boldsymbol{S}_{q, k}$ by (\ref{eq:S failure}) and (\ref{eq:S quality})\\
    Calculate ${ }^i \boldsymbol{P}_k$ by (\ref{eq:S variation})\\
    Construct ${ }^i \boldsymbol{W}_k$ by (\ref{eq:S weighing})\\
  }
  Update $\tilde{\boldsymbol{A}} = \bigoplus_{k=1}^{T_w} \tilde{\boldsymbol{A}}_k$ with $\tilde{\boldsymbol{A}}_k = \left[ -\boldsymbol{A}_k, \boldsymbol{I}_6 \right]$\\
  Update $\tilde{\boldsymbol{B}}=\bigoplus_{k=1}^{T_w}\boldsymbol{B}_k$ with $\boldsymbol{B}_k$\\

  \For{$k=0$ \KwTo $T_w$}{
    Update ${}^i\boldsymbol{C}_k$ with $\boldsymbol{\rho}_k=\boldsymbol{r}^{\top}_{k} /\left\|\boldsymbol{r}_{k}\right\|_2, \boldsymbol{\beta}_k$\\
    Update $\boldsymbol{\tau}_k=\bigoplus_{n=1}^{6} \boldsymbol{t}_k$ with $\boldsymbol{t}_k=\left[t_k^0, t_k^1, \cdots, t_k^{k_t}\right]$\\
    Update block matrices:\\
    \begin{itemize}
      \item $\boldsymbol{E}_{\boldsymbol{x}} \leftarrow \boldsymbol{I}_{n_x}, \tilde{\boldsymbol{A}}, \tilde{\boldsymbol{C}}$\\
      \item $\boldsymbol{E}_{\boldsymbol{\alpha}}  \leftarrow \boldsymbol{I}_{n_x}, \tilde{\boldsymbol{B}}, \boldsymbol{I}_{n_y} $\\
      \item $\boldsymbol W = \bigoplus_{k=0}^{T_w} {}^p\boldsymbol{W}_k \oplus \bigoplus_{k=1}^{T_w} {}^1\boldsymbol{W}_k \oplus \bigoplus_{k=0}^{T_w} {}^r\boldsymbol{W}_k$\\
    \end{itemize}
  }

  Assemble $\boldsymbol{\alpha}$ with $\check{\boldsymbol{x}}_k, \boldsymbol{u}_k, {}^i\boldsymbol{y}_k$\\
  Construct $\boldsymbol{E_\tau}=\boldsymbol{E_x} \boldsymbol\tau$\\
  Solve $\hat{\boldsymbol{x}}_{\boldsymbol\tau}$ by (\ref{eq:alphaest})\\
  Construct $\hat{\boldsymbol{x}}=\boldsymbol{\tau} \hat{\boldsymbol{x}}_{\boldsymbol\tau}$\\
  Update posterior state estimate $\hat{\boldsymbol{x}}_k$\\
\end{algorithm}

The optimal estimates of $\partial \boldsymbol J / \partial \hat{\boldsymbol{x}} =0$ can be obtained directly $\hat{\boldsymbol{x}}=\left(\boldsymbol{E}_{\boldsymbol{x}}^{\top} \boldsymbol{W} \boldsymbol{E}_{\boldsymbol{x}}\right)^{-1} \boldsymbol{E}_{\boldsymbol{x}}^{\top} \boldsymbol{W} \boldsymbol{E}_{\boldsymbol{\alpha}} \boldsymbol{\alpha}$. However, direct inversion of high-dimensional matrices within the sliding window incurs substantial computational overhead. To mitigate this, a dimension reduction method based on polynomial approximation is adopted, as proposed in \cite{dongSRIBOEfficientResilient2022}. Thus, the augmented dimension-reduced estimator is expressed as:
\begin{equation}
  \hat{\boldsymbol{x}}_{\boldsymbol\tau}=(\boldsymbol{E_{\tau}}^{\top}\boldsymbol{W}\boldsymbol{E_{\tau}})^{-1}\boldsymbol{E_{\tau}}^{\top}\boldsymbol{W}\boldsymbol{E_{\alpha}}\boldsymbol{\alpha}
  \label{eq:alphaest}
\end{equation}
\noindent where $\hat{\boldsymbol{x}}=\boldsymbol{\tau} \hat{\boldsymbol{x}}_{\boldsymbol\tau}$ and $\boldsymbol{E_\tau}=\boldsymbol{E_x} \boldsymbol\tau$. In the approximation process, each $\boldsymbol{\tau}_k=\bigoplus_{n=1}^{6} \boldsymbol{t}_k$ represents six copies of matrix $\boldsymbol{t}_k$, where $\boldsymbol{t}_k=\left[t_k^0, t_k^1, \cdots, t_k^{k_t}\right]$ corresponds to the $k_t$-th order polynomial fitting with $t_k^n=\left(t_k-t_0\right)^{n}$. Consequently, the dimension of $\boldsymbol{E}_{\boldsymbol x}^{\top}\boldsymbol W \boldsymbol{E_x}$ is reduced from $6(T_w+1) \times 6(T_w+1)$ to $6(k_t+1) \times 6(k_t+1)$. For each estimation, the weight matrix $\boldsymbol{W}$ will be updated adaptively. The overall estimation of the extended sliding window filter considering dimension reduced process with adaptive confidence evaluation is illustrated in Algorithm \ref{alg:algorithm1}.

\begin{figure}[!t]\centering
	\includegraphics[width=8.5cm]{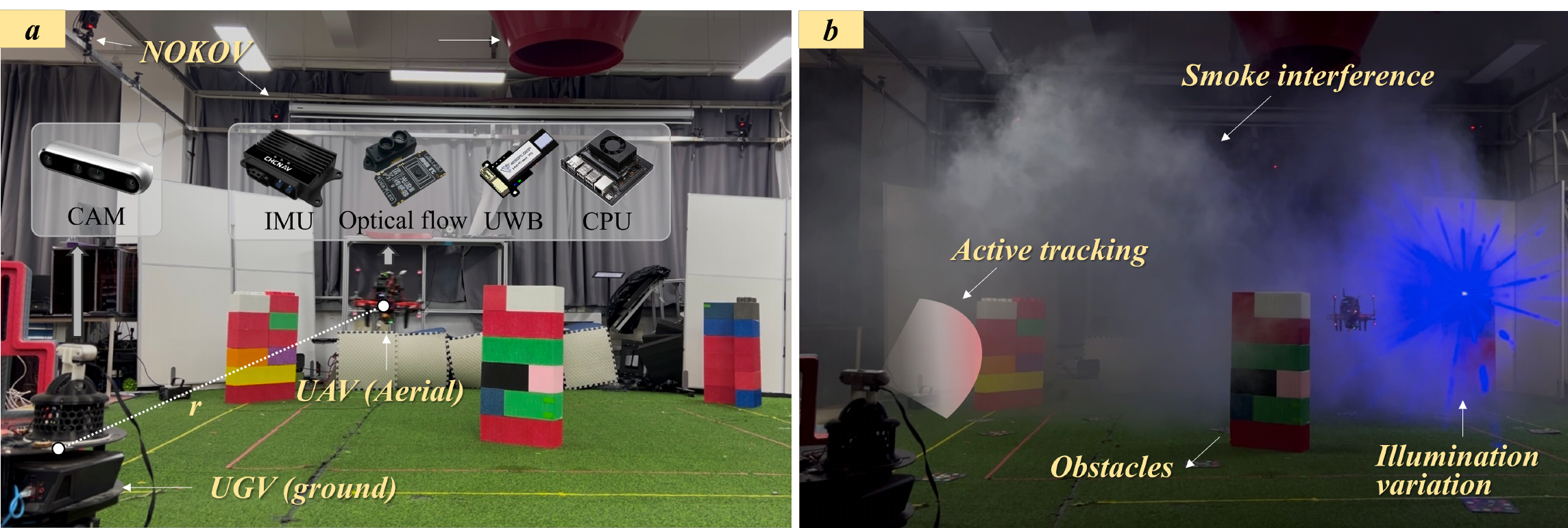}
	\caption{Experiment setup for the ground-aerial localization system. Subfigure (a) shows indoor Testbeds in clear scenario; subfigure (b) shows Testbeds in harsh scenario.}\label{fig:Testbeds}
\end{figure}

\section{Experiment}
\subsection{Experiment Setup}
To evaluate the validity, experiments are conducted in clear and harsh scenarios. In each experiment, the quadrotor is controlled by the open-source Pixhawk$^\circledR$firmware. An NVIDIA Jetson Xavier NX, together with an Intel Atom x7 (quad-core, 1.8 GHz), is used as the onboard computing platform. The IMU module CHCNAV CL-510 is utilized for acceleration measurements, while the NiMing v4 optical flow module is used for velocity. The Nooploop$^\circledR$ LinkTrack UWB radio is adopted to measure inter-agent distances. The IMU operates at 100 Hz, the UWB at 50 Hz, and the optical flow module at 25 Hz. The active vision mechanism is equipped with an Intel RealSense D455 camera running at 30 Hz. The mechanism is actuated by two orthogonally mounted SM40BL servo motors with integrated encoders, providing an angular resolution of 0.088$^{\circ}$. These servos are connected with the driver board via TTL-to-USB protocol. The ground mobile platform is the SSE1 model by EAI, featuring an STM32 control board and a dual-wheel differential-drive configuration. Its position feedback is obtained from the ground truth. The overall testbeds are illustrated in Fig. \ref{fig:Testbeds}.

The NOKOV motion capture system is used to obtain ground truth. The experiments include two typical scenarios: one with only obstacles and another with additional interference from smoke, varying illumination, and obstacles. An onboard controller is employed in all tests, with the aerial robot receiving positioning data from the real-time estimator. Besides, the ground vehicle uses ground truth for feedback. A local mesh network is established during flight using onboard sub-routers, enabling communication between the UAV (ROS master) and the UGV (ROS slave). Experiments are conducted under both relative static and dynamic conditions.

\begin{table}[t]
  \centering
  \caption{Comparisons of relative localization performance with different visual detection strategies.}
  \resizebox{\columnwidth}{!}{
  \begin{tabular}{clllllllc} 
  \toprule
  \multicolumn{1}{c}{\multirow{2}{*}{\textbf{Scen.}}} & \multicolumn{1}{c}{\multirow{2}{*}{\textbf{Method}}} & \multicolumn{3}{c}{\textbf{RMSE}($m$)} & \multicolumn{3}{c}{\textbf{MAE}($m$)} & \multicolumn{1}{c}{\multirow{2}{*}{$\eta(\%)$}}\\
  \cmidrule{3-8}
  \multicolumn{2}{c}{} & \multicolumn{1}{c}{$x$} & \multicolumn{1}{c}{$y$} & \multicolumn{1}{c}{$z$} & \multicolumn{1}{c}{$x$} & \multicolumn{1}{c}{$y$} & \multicolumn{1}{c}{$z$} & \\
  \midrule
  \multirow{3}{*}{\textbf{Clear}} & \textbf{F-yolo} & 0.110 & 0.106 & 0.030 & 0.093 & 0.081 & 0.025 & 20.102 \\
   & \textbf{F-pnp} & 0.055 & 0.069 & $\textbf{0.010}$ & 0.048 & 0.058 & $\textbf{0.007}$ & 16.606 \\
   & \textbf{A-pnp} & $\textbf{0.041}$ & $\textbf{0.052}$ & 0.011 & $\textbf{0.035}$ & $\textbf{0.044}$ & 0.008 & $\textbf{5.005}$ \\
  \midrule
  \multirow{3}{*}{\textbf{Harsh}} & \textbf{F-yolo} & 0.115 & 0.172 & 0.015 & 0.129 & 0.142 & 0.011 & 44.293 \\
   & \textbf{F-pnp} & 0.056 & 0.098 & $\textbf{0.012}$ & 0.045 & 0.073 & $\textbf{0.010}$ & 29.724 \\
   & \textbf{A-pnp} & $\textbf{0.041}$ & $\textbf{0.063}$ & 0.014 & $\textbf{0.034}$ & $\textbf{0.053}$ & $\textbf{0.010}$ & $\textbf{12.008}$ \\
  \toprule
  \end{tabular}}
  \label{tb:active vision}
\end{table}

\begin{figure}[!t]\centering
	\includegraphics[width=8.5cm]{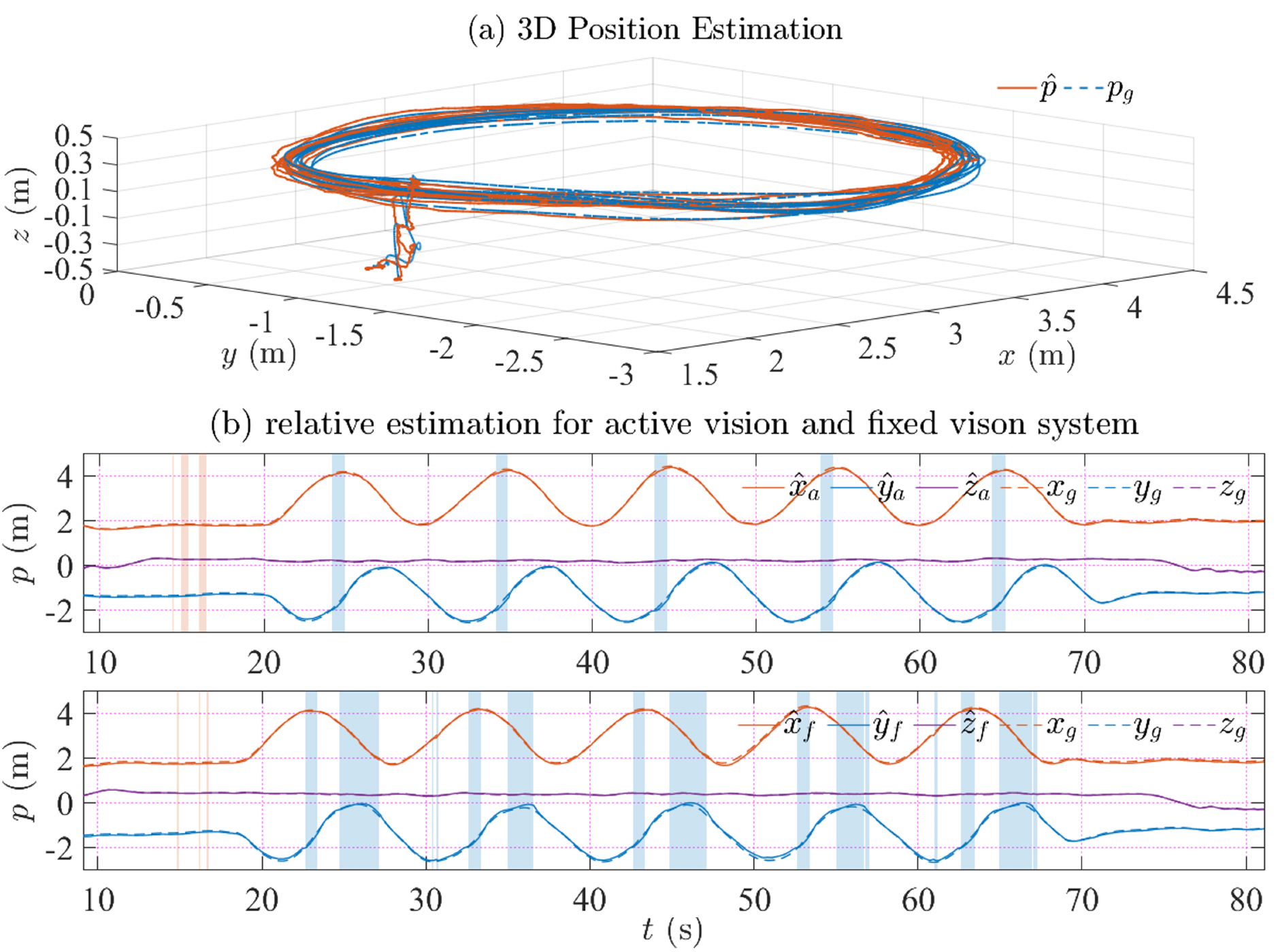}
	\caption{Comparison of relative localization for different visual detection strategies. Subfigure (a) illustrates the 3D estimated trajectory using the proposed method in a clear scenario. Subfigure (b) compares the proposed active-view based relative localization with a fixed-view system under the same conditions. The shaded regions indicate the time intervals of data loss.}
  \label{fg:visual loss}
\end{figure}

\subsection{Effectiveness for active estimation}
To validate the effectiveness of the proposed active vision mechanism, experiments were conducted using three visual detection strategies: fixed-view YOLO detection (F-yolo), fixed-view PnP (F-pnp), and active-view PnP (A-pnp). During experiments, the UGV remained stationary, while the UAV followed a circular trajectory with a radius of 1 m at a speed of 0.6 m/s. The flight altitude was set to 0.5 m in a clear environment and 0.7 m under harsh conditions with smoke and lighting interference. Above all tests, the online estimator achieves an average of 22.5 ms per iteration, while its update cycle is set to 25 Hz (40 ms) to match the controller's frequency. This ensures proper synchronization and avoids noticeable delays.

As shown in Table \ref{tb:active vision}, the root mean square error (RMSE) and mean absolute error (MAE) were evaluated along three axes. Fig. \ref{fg:visual loss} illustrates the estimated trajectories of the fixed-view and active-view methods in clear scenarios. The subscripts \( a \), \( f \), and \( g \) denote the active-view, fixed-view method, and ground truth, respectively. The orange-shaded regions indicate optical flow loss, while the blue ones represent visual loss.

It is worth noting that the proposed active-view PnP method consistently achieves high estimation accuracy, while the fixed-view PnP method performs slightly better along the z-axis. This can be attributed to the UAV maintaining an almost constant altitude during flight, resulting in small height variation. Additionally, minor differences in smoke dispersion across flight trials may have contributed to the slight improvement observed. Overall, the experimental results demonstrate that the proposed method provides superior robustness under smoke and illumination disturbances. Specifically, under harsh environmental conditions, the absolute trajectory error (ATE) is reduced by $67.0\%$ and $32.7\%$ compared with the F-yolo and F-pnp methods, respectively. Furthermore, the active vision mechanism effectively maintains continuous target tracking, reducing visual loss by $32.3\%$ and $17.7\%$.

\begin{figure}[!t]\centering
	\includegraphics[width=8.5cm]{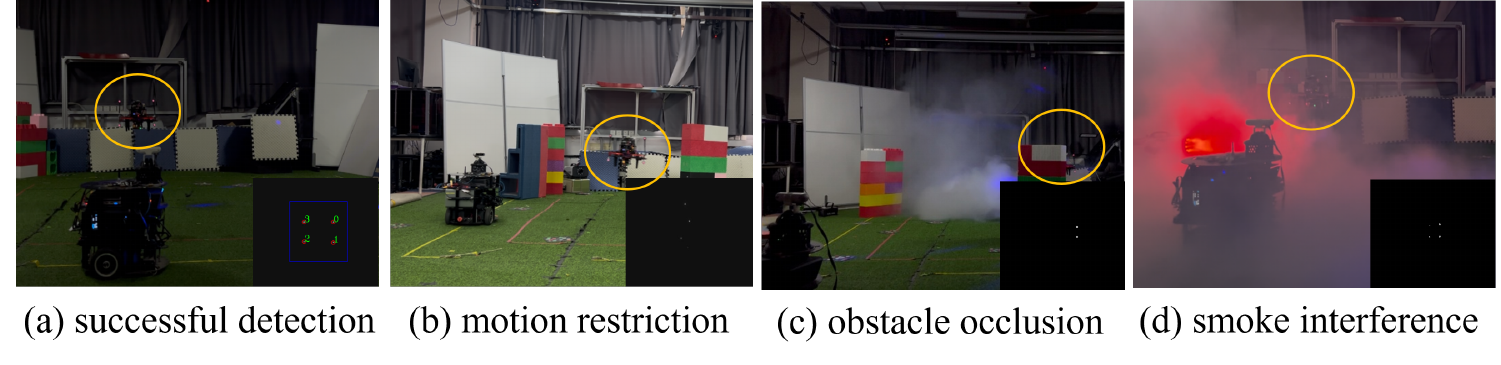}
	\caption{The typical visual detection failures for different trails and their corresponding experimental scenarios. The first-person view from the infrared camera is displayed in the bottom right corner of the figure.}
  \label{fig:M_h_visual}
\end{figure}

\begin{table}[t]
  \centering
  \caption{Estimation RMSE and MAE for diffenent trials.}
  \resizebox{\columnwidth}{!}{
  \begin{tabular}{cllllllc} 
  \toprule
  \multicolumn{1}{c}{\multirow{2}{*}{\textbf{Scen.}}} & \multicolumn{3}{c}{\textbf{RMSE} ($m$)} & \multicolumn{3}{c}{\textbf{MAE} ($m$)} & \multicolumn{1}{c}{$|\epsilon_{\max}|$ ($m$)}\\
  \cmidrule{2-8}
  \multicolumn{1}{c}{} & \multicolumn{1}{c}{$x$} & \multicolumn{1}{c}{$y$} & \multicolumn{1}{c}{$z$} &\multicolumn{1}{c}{$x$} & \multicolumn{1}{c}{$y$} & \multicolumn{1}{c}{$z$} & \multicolumn{1}{c}{$p$}  \\
  \midrule
  $S_1$ & 0.041 & 0.052 & 0.011 & 0.035 & 0.044 & 0.008 & 0.192\\
  $S_2$ & 0.041 & 0.063 & 0.014 & 0.034 & 0.053 & 0.010 & 0.194\\
  $L_1$ & 0.068 & 0.087 & 0.010 & 0.050 & 0.061 & 0.007 & 0.396\\
  $L_2$ & 0.052 & 0.103 & 0.018 & 0.040 & 0.076 & 0.014 & 0.320\\
  $M_1$ & 0.028 & 0.061 & 0.011 & 0.024 & 0.045 & 0.009 & 0.155\\
  $M_2$ & 0.083 & 0.073 & 0.026 & 0.056 & 0.058 & 0.017 & 0.406\\
  \toprule
  \end{tabular}}
  \label{tb:adaptive}
\end{table}

\subsection{Robustness for adaptive estimation}
To verify the robustness of A2SVIR, extensive evaluations were conducted in challenging scenarios. In addition to environments with smoke interference, illumination changes, and obstacle occlusion, diverse conditions, including prolonged visual loss, relative motion, and cluttered or large-scale outdoor localization tests, are implemented. Typical visual detection failures in corresponding scenarios are shown in Fig. \ref{fig:M_h_visual}. Specifically, indoor experiments are carried out in both fixed and dynamic anchor scenarios. In the fixed-anchor scenarios (stationary UGV), tests are conducted under three conditions: clear environment ($S_1$), harsh environment ($S_2$), and prolonged visual loss ($L_1,L_2$). The dynamic-anchor setup (moving UGV) includes two cases: in case ($M_1$), the UAV maintains a stationary position relative to the UGV as it moves back and forth in a degraded environment, while in the collaborative motion case ($M_2$), the UGV and UAV follow different trajectories in a clear environment. The estimation results are summarized in Table \ref{tb:adaptive}. The dataset used for the estimation process is also released as open source on a GitHub repository, providing both ROS bags and the corresponding ROS-node implementation at \url{https://github.com/scarlettchen618/dataset_for_a2visr.git}.

\begin{figure}[!t]\centering
	\includegraphics[width=8.5cm]{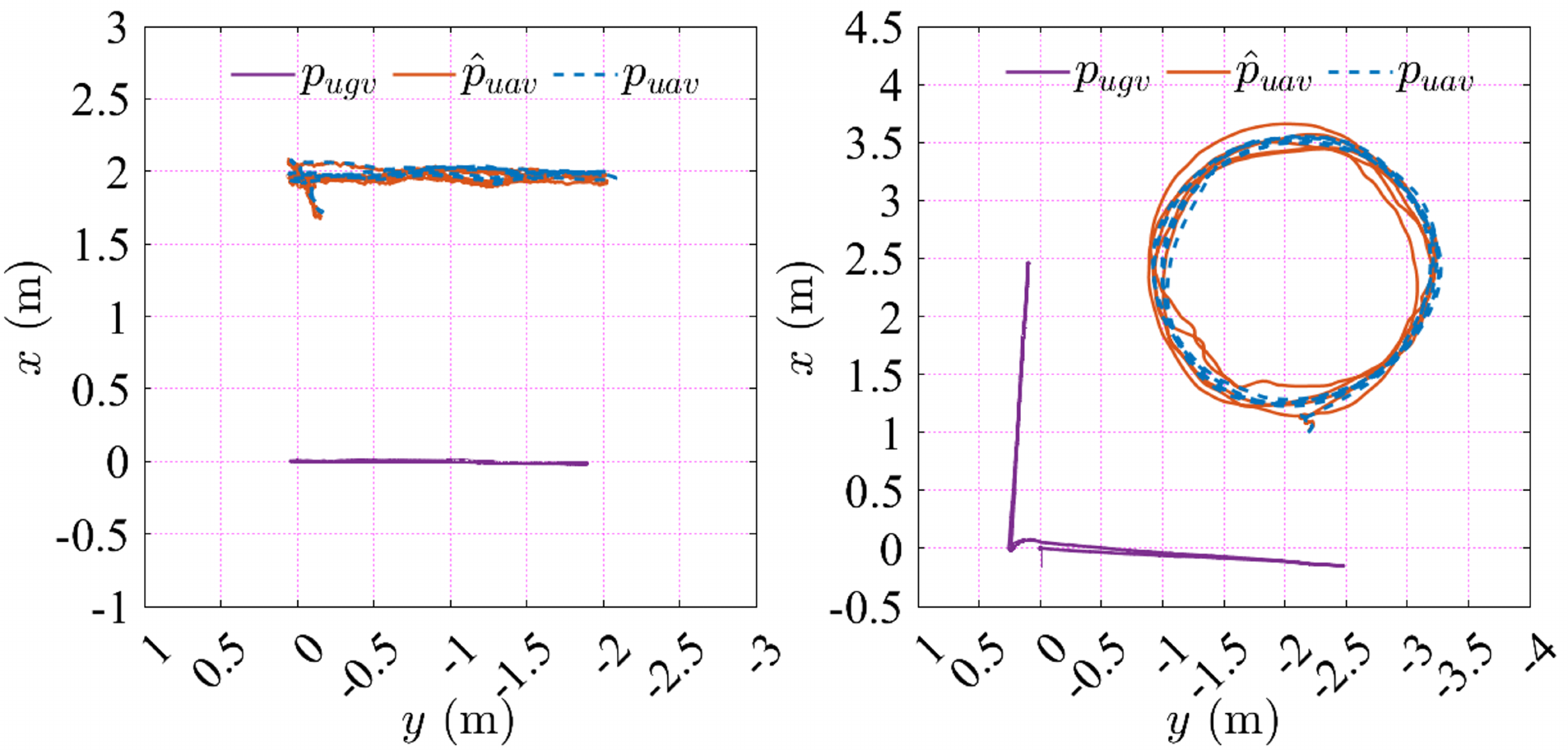}
	\caption{The top view of relative localization in the ground-aerial cooperation system. Subfigure (a) illustrates the relative hover motion. Subfigure (b) depicts the relative motion with different trajectories.}\label{fig:motion}
\end{figure}

\subsubsection{Robustness for dynamic anchor}
Before real flight, we record a dataset based on the current experimental setup for parameter calibration and fine-tune additional parameters using a trial-and-error approach. Subsequently, the estimation and control processes are conducted entirely onboard. Specifically, the aerial drag coefficient is set as $\boldsymbol{\mu}=\mathrm{diag}(0.2,0.2,0.2)$, with a window size $T_w = 8$ and a polynomial fitting order of $k_t = 3$ for online estimation. The initial weighing matrix is set as $^p \boldsymbol{W}_0=\mathrm{diag}(0.1,0.1,0.1)$, while $^j \boldsymbol{W}_0 (j=1,2,\cdots,5)$ are each set as identity matrices.

As shown in Fig. \ref{fig:motion}, the top-view trajectories are plotted in the system's global coordinate frame. The results of the dynamic anchor experiments ($M_1$ and $M_2$) demonstrate that the proposed A2SVIR system can provide continuous and stable localization for the aerial robot under relative motion conditions. Whether maintaining a static relative position or following different motion trajectories, the system achieves high-precision and robust localization, with average RMSE and MAE of 0.092 m and 0.070 m, respectively.

\subsubsection{Robustness for prolonged visual loss}
To further assess the stability of the proposed method under prolonged visual loss, we conducted experiments $L_1$ and $L_2$, where the visual detection weight factor was manually set to zero during the 30 s-40 s and 50 s-60 s intervals for $L_1$, and during the 15 s-30 s interval for $L_2$ to simulate extended vision failure. Throughout this period, the active vision mechanism relied on estimation feedback to maintain continuous target tracking. The results demonstrate that the system effectively handles sudden and prolonged visual loss while maintaining an estimation error of approximately 0.010 m, verifying its robustness.

\begin{figure}[!t]\centering
	\includegraphics[width=8.5cm]{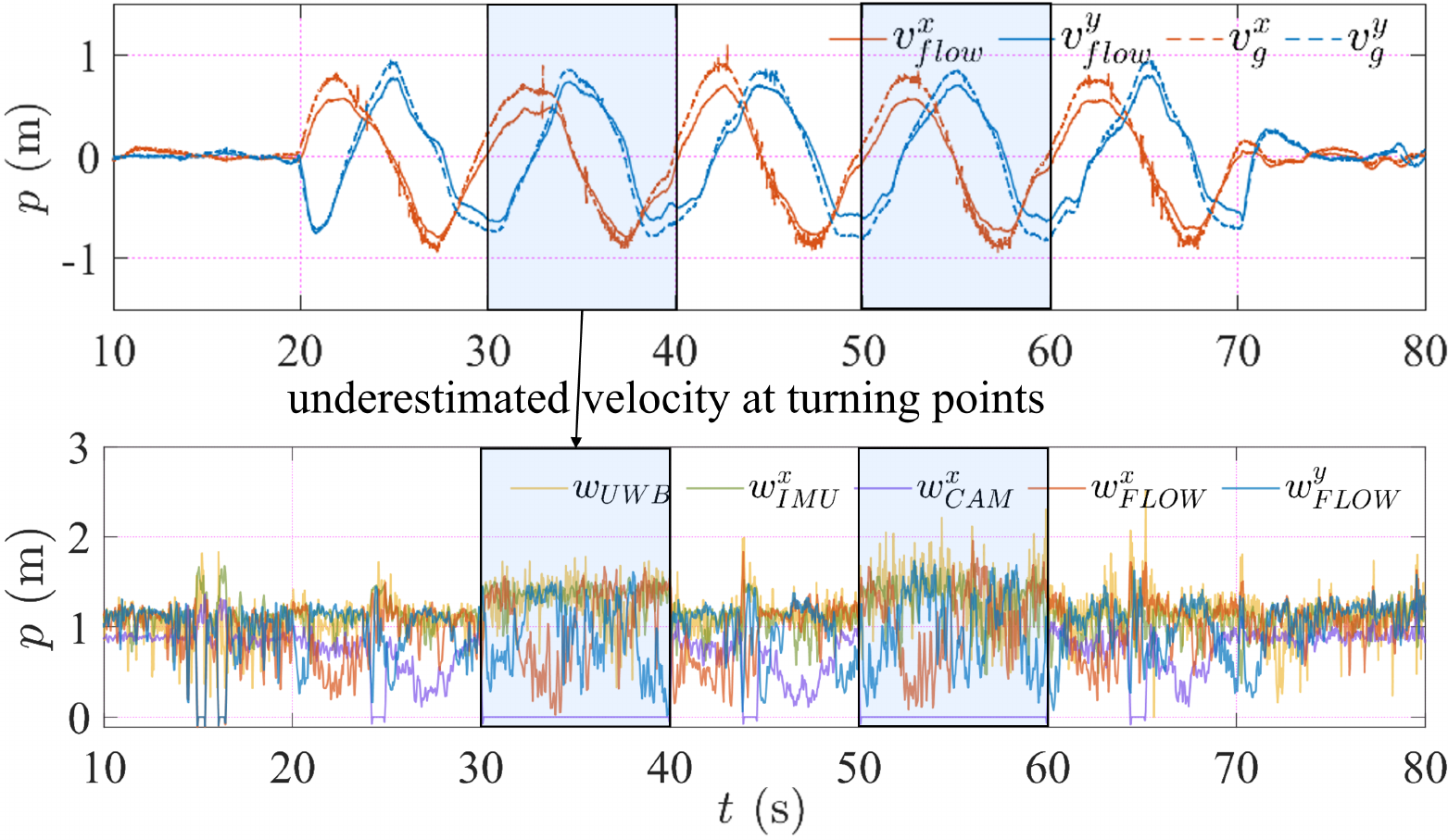}
	\caption{The optical flow velocity measurements and adaptive weighing parameter adjustment for trail $L_1$ with long time visual loss.}
  \label{fg:long time lost}
\end{figure}

A detailed analysis of trajectory $L_1$, including adaptive confidence evaluation and a comparison between optical flow measurements and velocity ground truth, is presented in Fig. \ref{fg:long time lost}. Results indicate that the optical flow sensor tends to underestimate during peak velocity transitions. The adaptive sliding confidence evaluation effectively detects these variations and dynamically reduces the weight of optical flow feedback, enhancing the estimation accuracy.

\begin{figure}[!t]\centering
	\includegraphics[width=8.5cm]{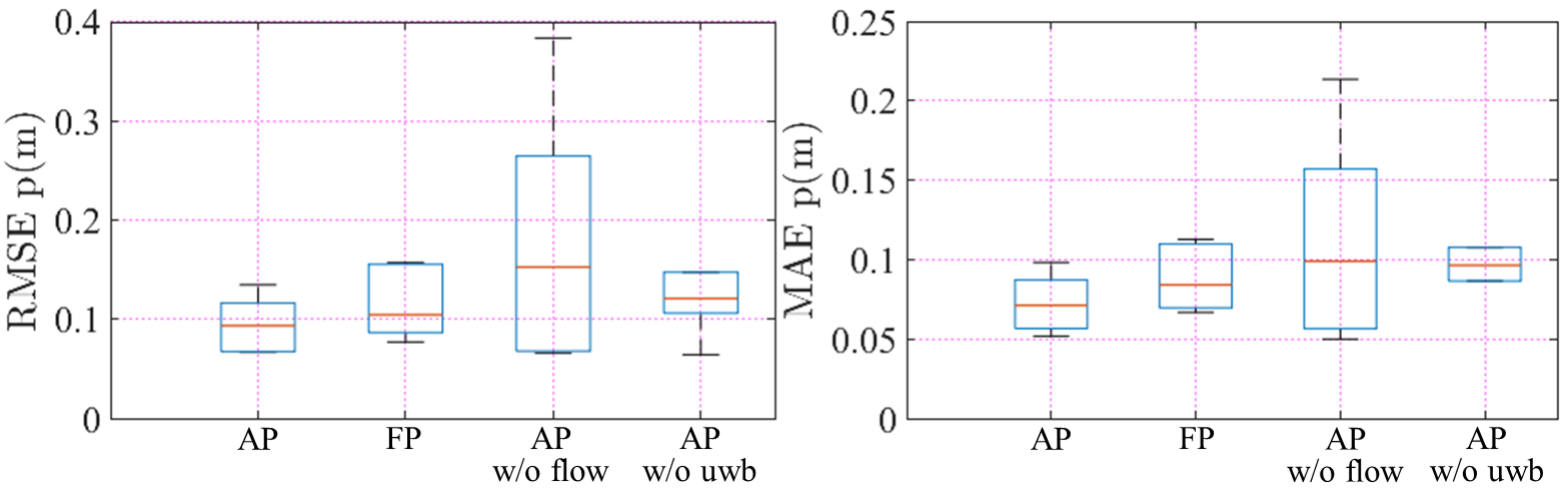}
	\caption{Comparison of RMSE and MAE results obtained using different estimation methods. The solid line represents the medians, while the blue dotted line represents the mean of RMSE and MAE.}
  \label{fg:comparison for adaptive method}
\end{figure}

Furthermore, we conducted a comparative study using fixed-weight parameters. By processing recorded rosbag data offline, we performed only fault detection while estimating with the fixed initial weight matrix. Additional tests were carried out under the simulation, including optical flow loss and UWB failure. The aerial drag coefficient is set as $\boldsymbol{\mu}=\mathrm{diag}(1.2,0.2,1.2)$ for simulations without optical flow or UWB. The box plot of RMSE and MAE is shown in Fig. \ref{fg:comparison for adaptive method}. The results indicate that, compared to fixed-parameter methods, the adaptive sliding confidence evaluation strategy demonstrates superior adaptability to sudden sensor failures. Moreover, due to the higher measurement accuracy of optical sensors, the absence of optical flow data has a more pronounced impact on overall system estimation accuracy.

\begin{figure}[t]\centering
	\includegraphics[width=8.5cm]{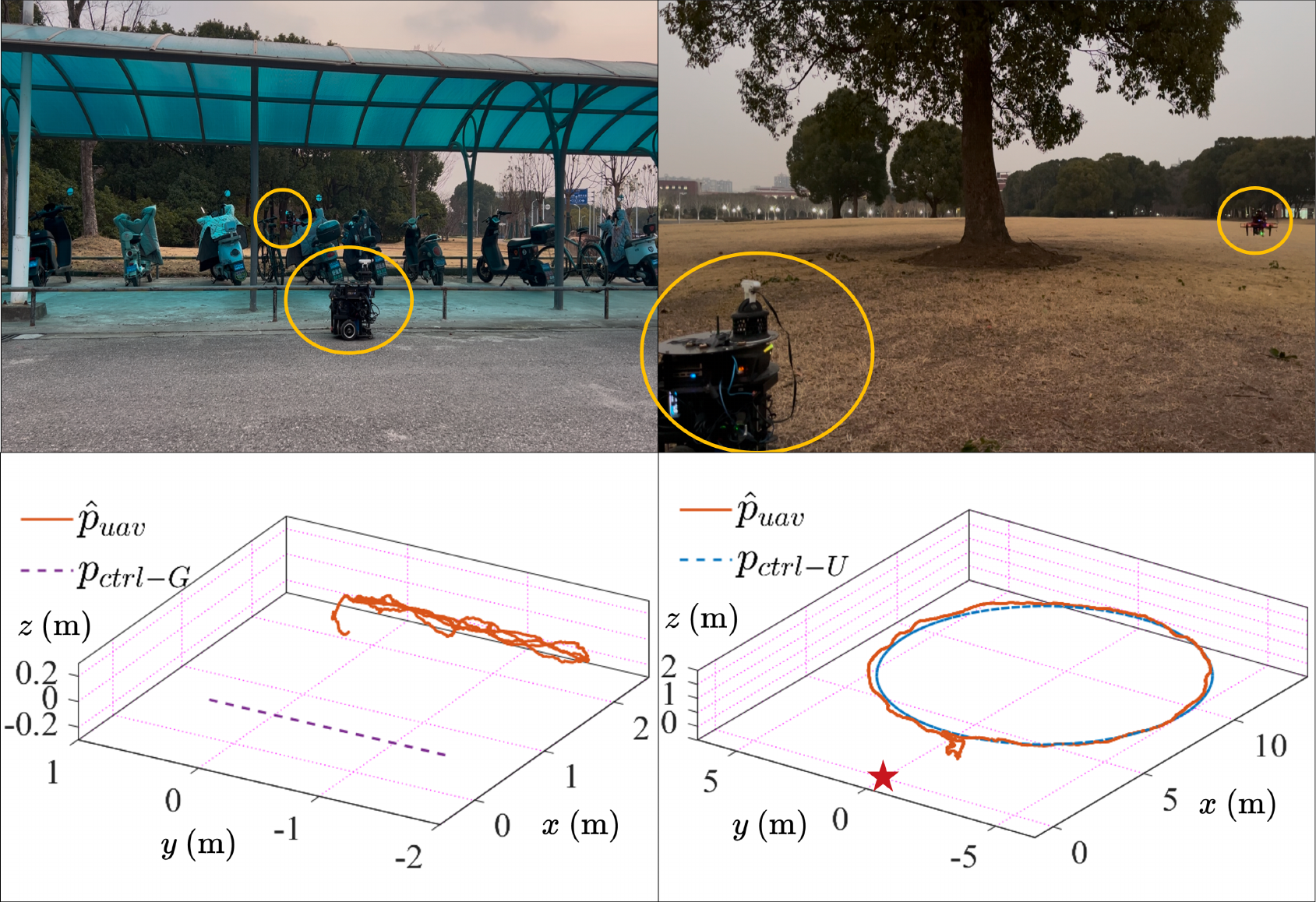}
	\caption{Outdoor experiments in cluttered and long-range scenarios.}
  \label{fg:outdoor}
\end{figure}

\begin{table}[t]
  \centering
  \caption{ATE ($m$) for diffenent methods in typical trials.}
  \resizebox{\columnwidth}{!}{
  \begin{tabular}{lcccc} 
  \toprule
  \textbf{Typical trials} & \textbf{SWF} \cite{9829196} & \textbf{RLS} \cite{10226597} & \textbf{KF based} \cite{8954658} & \textbf{Proposed} \\  
  \midrule
  Stationary anchor ($S_2$) & $0.131$ & $0.174$ & $0.150$ & $\textbf{0.077}$ \\
  Prolonged visual loss ($L_{1}$)  & $0.291$ & $0.265$ &$0.255$ & $\textbf{0.117}$ \\
  Dynamic anchor ($M_{2}$) & $0.361$ & $0.395$ & $0.340$ & $\textbf{0.135}$ \\
  \toprule
  \end{tabular}}
  \label{tb:compare}
  \end{table}

\subsubsection{Robustness for long-range scenario}
As shown in Fig. \ref{fg:outdoor}, outdoor experiments were conducted to evaluate the system's performance in cluttered environments under relatively static conditions and in long-range scenarios. The estimations are referenced to the global coordinate frame, which is derived from the initial pose of the ground vehicle. The estimated trajectories depicted in Fig. \ref{fg:outdoor} correspond to the actual camera viewpoints recorded during the experiments. In the relative static case, tests were carried out in a bicycle shed. The UGV moved back and forth in a straight line at constant speed using open-loop control, while the UAV maintained relative hovering via onboard control. For the long-range experiment, conducted at dusk under low-light conditions, the UGV remained stationary as the UAV followed a circular trajectory with a 5.5m radius under onboard control.

Since MCS cannot be deployed in outdoor environments and quantitative accuracy verification has been extensively conducted indoors, the outdoor experiments were primarily designed for qualitative validation. As shown in Fig. \ref{fg:outdoor}, the system successfully performs localization at relative distances up to 12 m. Compared to active LED-based methods in \cite{8967660,7487248}, which are limited to an operational range of approximately 5 m, the proposed method demonstrates superior feasibility in long-range and complex environments, further validating its robustness and applicability under challenging conditions.

\subsubsection{Effectiveness for performance comparison}
Furthermore, we compared our proposed method with state-of-the-art estimators to validate its high accuracy and robustness. The results against general Sliding Window Filter (SWF) \cite{9829196}, Recursive Least Squares (RLS) \cite{10226597}, and Kalman Filter (KF) based estimator \cite{8954658} are summarized in Table \ref{tb:compare}, evaluating the estimation absolute trajectory error (ATE). As shown, our method consistently achieves the highest estimation accuracy. Overall, these results demonstrate the superior performance of the proposed approach.

\section{Conclusion}
In this paper, we propose an active and adaptive ground-aerial localization framework that leverages active visual feedback, single-range, and inertial fusion. The framework is validated through extensive experiments under challenging conditions. Results demonstrate that the active vision subsystem effectively enhances the target tracking performance, while the reformulated dimension-reduced estimator with adaptive sliding confidence evaluation effectively assesses sudden sensor failures and degradations, adjusting confidence levels accordingly. The proposed A2SVIR framework achieves an average trajectory RMSE of 0.092 m across various scenarios, with a notably low RMSE of 0.068 m in clear environments. Furthermore, qualitative evaluations confirm its effectiveness in estimating relative motion in cluttered scenarios and performing large-scale localization outdoors, demonstrating the system's robustness and resilience.

In the future, we will further explore mutual and active observation for multiple ground monitors to enhance the practicality for formation control and collaborative mapping.

\vspace{-10mm} 

\begin{IEEEbiography}[{\includegraphics[width=1in,height=1.25in,clip,keepaspectratio]{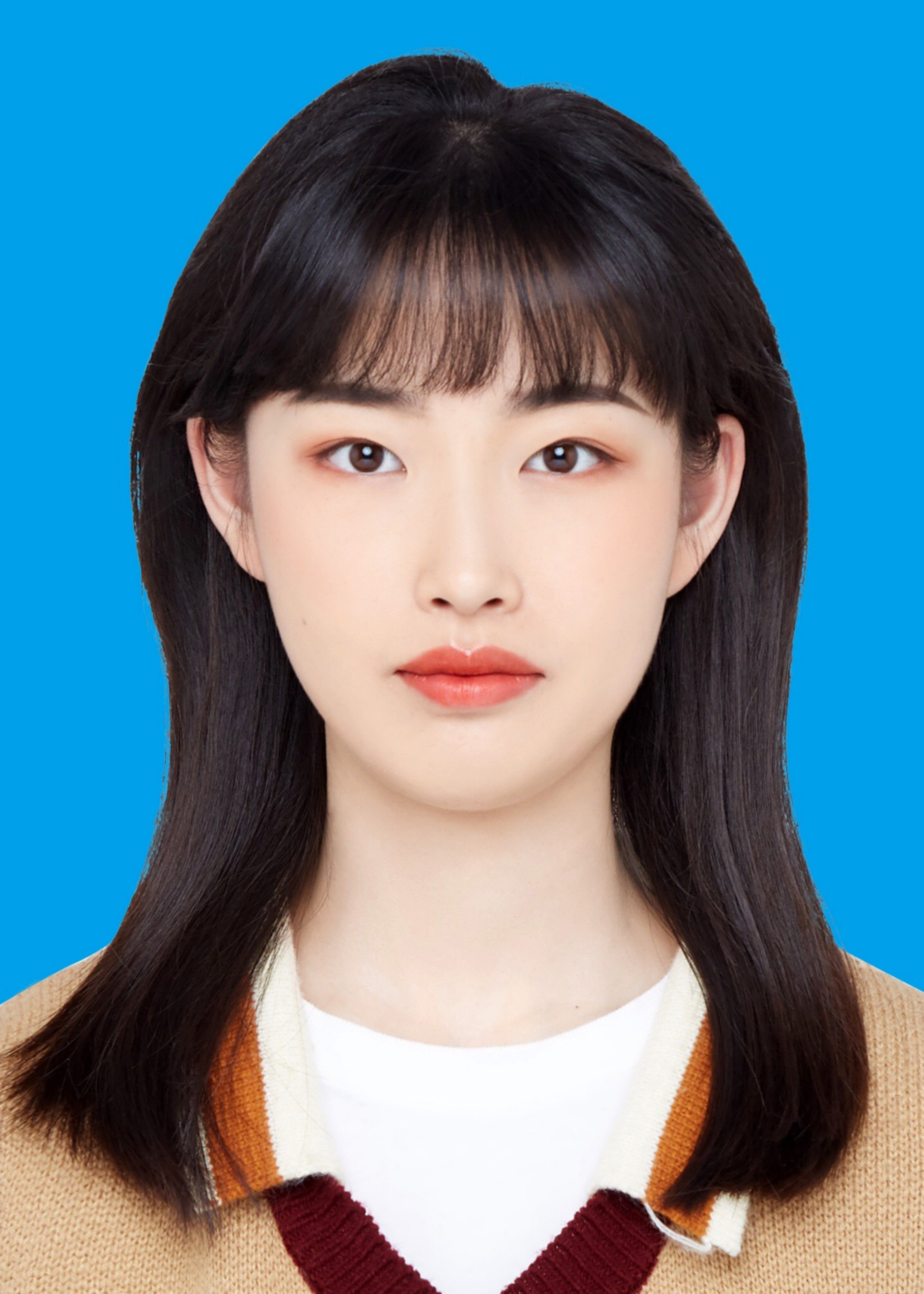}}]
	{Sijia Chen} received the B.S. degree in mechanical design manufacture and automation from the University of Electronic Science and Technology of China, Sichuan, China, in 2022. She is currently a Ph.D. candidate with the State Key Laboratory of Mechanical System and Vibration, School of Mechanical Engineering, Shanghai Jiao Tong University. Her research interests include state estimation and intelligent control of unmanned systems.
\end{IEEEbiography}
	
\vspace{-10mm} 

\begin{IEEEbiography}[{\includegraphics[width=1in,height=1.25in,clip,keepaspectratio]{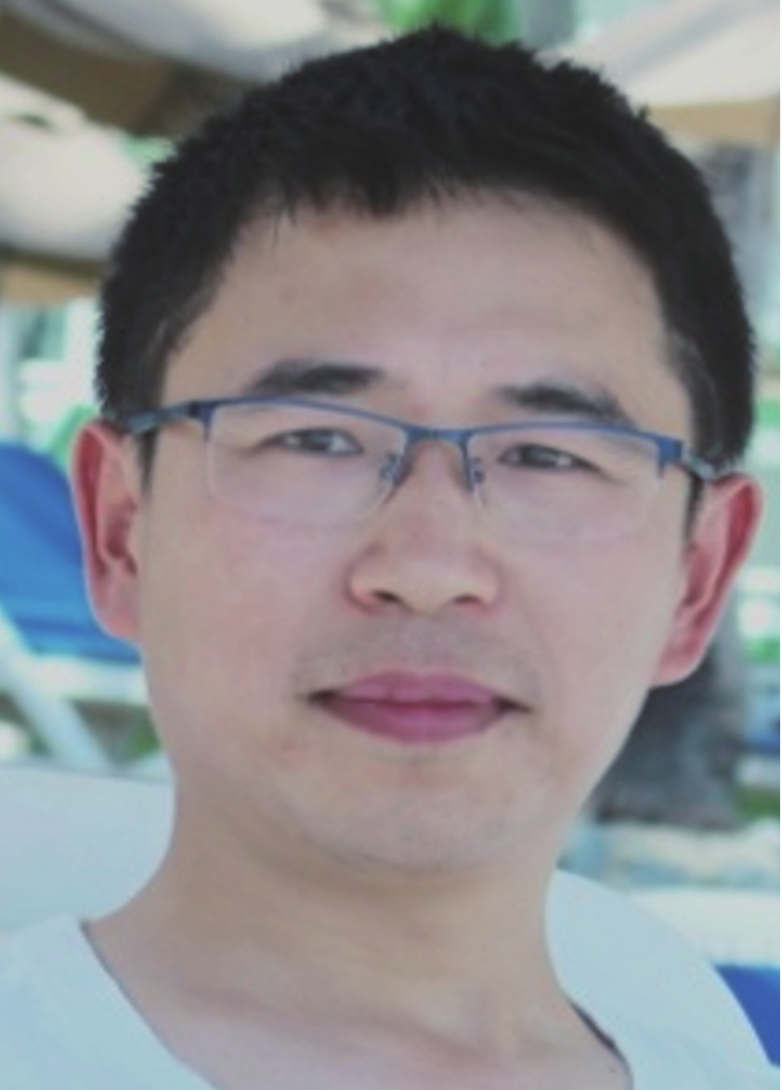}}]
	{Wei Dong} received the B.S. degree and Ph.D. degree in mechanical engineering from Shanghai Jiao Tong University, Shanghai, China, in 2009 and 2015, respectively. He is currently an associate professor in the Robotic Institute, School of Mechanical Engineering, Shanghai Jiao Tong University. For years, his research group was champions in several national-wide autonomous navigation competitions of unmanned aerial vehicles in China. In 2022, he was selected into the Shanghai Rising-Star Program for distinguished young scientists. His research interests include cooperation, perception and agile control of unmanned systems.
\end{IEEEbiography}

\end{document}